\documentclass[letterpaper, 10 pt, conference]{ieeeconf}   

\IEEEoverridecommandlockouts
\usepackage{cite}
\usepackage{bm}
\usepackage{amsmath,amssymb,amsfonts}
\usepackage{algorithmic}
\usepackage{graphicx}
\usepackage{textcomp}
\usepackage[dvipsnames]{xcolor}
\usepackage[colorlinks,citecolor=red,urlcolor=black,bookmarks=false,hypertexnames=true]{hyperref}
\def\BibTeX{{\rm B\kern-.05em{\sc i\kern-.025em b}\kern-.08em
    T\kern-.1667em\lower.7ex\hbox{E}\kern-.125emX}}

\def\isconfidential{0}

\if\isconfidential1
\usepackage{fancyhdr}
\fi
\newcommand{\makeconfidential}{
	\if\isconfidential1
	\fancyhead[C]{{\bf MERL CONFIDENTIAL}. $\copyright$MERL, $2024$. }
	\fancyfoot[C]{{\bf MERL CONFIDENTIAL}. $\copyright$MERL, $2024$}
	\thispagestyle{fancy}
	\pagestyle{fancy}
	\else
	\thispagestyle{empty}
	\pagestyle{empty}
	\fi
}

\begin{document}

\newcommand{\bolded}[1]{\bm{#1}}
\newcommand{\I}{\boldsymbol I}
\newcommand{\0}{\boldsymbol 0}
\newcommand{\1}{\boldsymbol 1}
\newcommand{\Jac}{\boldsymbol{J}}
\newcommand{\prob}{{\rm Pr}}

\title{\LARGE \bf Simultaneous Collision Detection and Force Estimation for Dynamic Quadrupedal Locomotion\\
}

\author{
\large Ziyi Zhou$^{1}$, \large Stefano Di Cairano$^{1}$, \large Yebin Wang$^{1}$, \large Karl Berntorp$^{1}$
\thanks{$^{1}$Mitsubishi Electric Research Laboratories (MERL), 02139 Cambridge,
MA, USA. {\tt\small karl.o.berntorp@ieee.org.}}
}

\maketitle
\makeconfidential

\begin{abstract}
In this paper we address the simultaneous collision detection and force estimation problem for quadrupedal locomotion using joint encoder information and the robot dynamics only. We design an interacting multiple-model Kalman filter (IMM-KF) that estimates the external force exerted on the robot and multiple possible contact modes. The method is invariant to any gait pattern design. Our approach leverages pseudo-measurement information of the external forces based on the robot dynamics and encoder information. Based on the estimated contact mode and external force, we design a reflex motion and an admittance controller for the swing leg to avoid collisions by adjusting the leg's reference motion. Additionally, we implement a force-adaptive model predictive controller to enhance balancing. Simulation ablatation studies and experiments show the efficacy of the approach.
\end{abstract}

\section{Introduction}
Quadrupedal locomotion has made significant progress over the past few decades, demonstrating exceptional dynamic behavior and robustness against external disturbances \cite{di2018dynamic,kim2019highly,zhu2021terrain}. To enhance the capabilities of legged robots on more challenging terrains, such as stairs or rocks, successful approaches \cite{kim2020vision,agrawal2022vision,jenelten2020perceptive,grandia2023perceptive} have integrated terrain information into planning or estimation \cite{varin2020constrained} modules to intentionally reduce the risk of tripping. However, due to the limited bandwidth of sensor data and potential tracking errors, reliable descriptions of surrounding terrains are not always available, particularly for legged robots moving at high speeds. This limitation underscores the importance of of robust collision detection and contact force estimation in dynamic locomotion independent on external sensors, allowing rapid reactions to unexpected contact events.

Independent of additional sensors, which are expensive and prone to damage, proprioceptive collision detection for rigid robots has garnered significant attention \cite{haddadin2017robot}. For legged locomotion, collision detection presents a greater challenge due to the frequent intentional contacts inherent in movement. A critical problem is distinguishing between different contact modes, such as regular stance mode and unexpected collisions. Heuristics-based approaches have been successfully deployed on quadruped robots. For example, in \cite{focchi2013local}, unexpected collisions are detected based on prior knowledge of the locomotion gait and the shape of friction cones, following a rough estimation of external contact forces. Additionally, \cite{hwangbo2016probabilistic} explicitly builds a Hidden Markov Model, which is later extended to contact and slip detection in \cite{jenelten2019dynamic}. However, the aforementioned works primarily focus on contact or collision detection, without accurately estimating the external contact force or wrench for further use.

The Momentum-Based Observer (MBO) \cite{de2006collision} offers accurate estimation of external torques without requiring joint acceleration measurements, which can be further utilized to compute external contact wrenches when the contact location is known. MBO has been effectively applied in various aspects of dynamic locomotion, including contact detection (stance or swing) \cite{bledt2018contact,cai2023predefined,cha2023implementation}, collision detection \cite{yim2023proprioception,van2022collision}, and disturbance rejection \cite{morlando2021whole,jenelten2022tamols}. In particular, prior work on contact or collision detection often employs a hierarchical pipeline, first estimating external forces and then using post-processing techniques, such as thresholding, to determine whether contact has occurred. While a robust thresholding method was proposed in \cite{van2022collision} to account for model uncertainties, developing a robust criterion that considers multiple contact modes beyond binary contact detection remains challenging. To achieve reliable performance, additional filtering and heuristics, such as scheduled contact timing from the controller design, are typically necessary \cite{bledt2018contact,cha2023implementation}. To the best of our knowledge, no existing collision detection approach simultaneously estimates multiple contact modes and the external force while remaining invariant to any gait pattern design.

To this end, we propose an approach based on the Interacting Multiple-Model Kalman Filter (IMM-KF) \cite{blom1988interacting} to simultaneously estimate external forces and contact modes, including swing, stance, and collision, only using encoder information and the robot dynamics. Each KF within the IMM-KF is tailored to a specific contact mode, where the dynamics of generalized momentum (GM) are integrated with a disturbance observer for contact forces \cite{wahrburg2015cartesian}. In the measurement model, we incorporate pseudo-measurements for the external forces derived from the robot's dynamics. The IMM-KF then fuses these contact modes and determines the probability of each mode occurring. Upon estimating the contact mode and external forces, we implement an admittance control law to respond effectively to detected collisions. In addition, the estimated contact force is sent to a model predictive controller (MPC) \cite{di2018dynamic,schperberg2022auto} for disturbance rejection, ensuring robust and adaptive locomotion with minimal sensing capabilities.

The main contributions of this work are:
\begin{itemize}
    \item We develop an estimator based on the IMM-KF that is independent of any specific controller design. By integrating GM with desired ground reaction forces, we provide enhanced feedback for distinguishing between contact and collision modes.
    
    \item Leveraging the collision detection capability, we design a reflex motion response. Additionally, the estimated forces are fed into an MPC module for improved balance control.

    \item We validate our approach through ablation studies and comparisons with other baselines in both simulation and hardware environments.
\end{itemize}

\section{Preliminaries}
This section contains brief descriptions of momentum-based observers and Kalman filters to facilitate understanding of the proposed approach.
\subsection{Momentum-Based Observer}
The robot dynamics for a quadrupedal robot with $n$ degrees of freedom (DoF) can be expressed as
\begin{equation}\label{eq:eom}
    \bolded{M}(\bolded{q})\ddot{\bolded{q}} + \bolded{C}(\bolded{q},\dot{\bolded{q}})\dot{\bolded{q}} + \bolded{g}(\bolded{q}) + \bolded{\tau}_f = \bolded{\tau}_{\text{ext}} + \bolded{\tau}_{m},
\end{equation}
where $\bolded{M} \in \mathbb{R}^{n \times n}$ is the mass matrix, $\bolded{C}(\bolded{q},\dot{\bolded{q}}) \in \mathbb{R}^{n \times n}$ is the Christoffel's symbol due to Coriolis and centrifugal terms, $\bolded{q}$ is the joint angle, $\bolded{g}$ is the gravity term, $\bolded{\tau}_f$ is the joint frictional torque, $\bolded{\tau}_m$ is the joint torque induced by the motors, and $\bolded{\tau}_{\text{ext}}$ is the external joint torque induced by the ground reaction forces. The external joint torque is
\begin{equation}
    \bolded{\tau}_{\text{ext}} = \sum_{j=1}^{n_f} \bolded{J}_{c_j}^T(\bolded{q})\bolded{\mathcal{F}}_{\text{ext},j},
\end{equation}
where $\bolded{J}_{c_j}(\bolded{q}) \in \mathbb{R}^{6 \times n}$ and $\bolded{\mathcal{F}}_{\text{ext},j}$ are the geometric contact Jacobian and external wrenches applied at the $j$th foot, respectively. $n_f$ is the number of feet of the robot. The external wrench includes both the external force $\bolded{f}_{\text{ext}}$ and moment $\bolded{m}_{\text{ext}}$, i.e.,
\begin{equation}
    \bolded{\mathcal{F}}_{\text{ext}} = \begin{bmatrix}
        \bolded{f}_{\text{ext}}\\
        \bolded{m}_{\text{ext}}
    \end{bmatrix} \in \mathbb{R}^6.
\end{equation}
The generalized momentum (GM) $\bolded{p}$ can be defined as
\begin{equation}\label{eq:gm}
    \bolded{p} = \bolded{M}\dot{\bolded{q}}
\end{equation}
where due to the skew-symmetry property of the robot dynamics,
\begin{equation}
    \dot{\bolded{M}}(\bolded{q}) = \bolded{C}(\bolded{q},\dot{\bolded{q}}) +
    \bolded{C}^{T}(\bolded{q},\dot{\bolded{q}}),
\end{equation}
the dynamics of the estimated GM $\hat{\bolded{p}}$ can be written as
\begin{equation}
    \dot{\hat{\bolded{p}}} = \bolded{\tau}_m - \bolded{\tau}_f + \bolded{C}^T\dot{\bolded{q}}-\bolded{g}+\bolded{\tau}_{\text{ext}}.
\end{equation}
Then, further filtering can be applied either using a first-order filter \cite{de2006collision} or a Kalman filter \cite{wahrburg2015cartesian}, with $\bolded{p} = \hat{\bolded{M}}\dot{\bolded{q}}$ as the measured GM and $\hat{\bolded{M}}$ as the estimated mass matrix.

\subsection{Single-Model Kalman Filter}
\label{ssec:single_kalman}
A Kalman filter is a recursive state estimator for a dynamical system of the form
\begin{subequations}
\begin{align} 
    \boldsymbol{x}_{t+1}
    &=
    \boldsymbol{A}_{t}
    \boldsymbol{x}_{t}
    +
    \boldsymbol{B}_{t}
    \boldsymbol{u}_{t}
    + \boldsymbol{w}_t,
    \\
    \boldsymbol{y}_{t} &= \boldsymbol{C}_{t}\boldsymbol{x}_{t}
    +
    \boldsymbol{v}_{t},
\end{align} 
\end{subequations}
where $\boldsymbol{x}_t\in\mathbf{R}^{n_x}$ is the state at time step $t$, $\boldsymbol{u}_t$ is the input, $\boldsymbol{y}_t\in\mathbf{R}^{n_y}$ is the output vector, $ \boldsymbol{w}_t \sim\mathcal{N}(\0,\boldsymbol{Q}_t)$ defines the process noise, and $ \boldsymbol{v}_t \sim\mathcal{N}(\0,\boldsymbol{R}_t)$ defines the measurement noise. 
At each recursion, the Kalman filter executes a prediction step and an update step~\cite{welch1995introduction}.
\begin{subequations}
\subsubsection{Prediction step}
In the prediction step, a model of the dynamical system's evolution is used to propagate and predict the first and second moments of the posterior distribution,
\begin{align}
        \hat{\boldsymbol{x}}_{t|t-1}
        &=
        {\boldsymbol{A}}_{t-1}
        \hat{\boldsymbol{x}}_{t-1|t-1}
        +
        {\boldsymbol{B}}_{t-1}
        {\boldsymbol{u}}_{t-1},
        \\
        {\boldsymbol{P}}_{t|t-1}
        &=
        {\boldsymbol{A}}_{t-1}
        {\boldsymbol{P}}_{t-1|t-1}
        {\boldsymbol{A}}_{t-1}^T
        +
        \boldsymbol{Q}_{t-1},
\end{align}
where $\hat{\boldsymbol{x}}_{t|t}$ and ${\boldsymbol{P}}_{t|t}$ are the mean and covariance of the distribution describing the state estimate at time step $t$, and $\hat{\boldsymbol{x}}_{t|t-1}$ and ${\boldsymbol{P}}_{t|t-1}$ are the predicted mean and covariance.

\subsubsection{Update step}
In the update step, the predicted mean and covariance are updated using sensor measurements:
    \begin{align}
        \tilde{\boldsymbol{y}}_t
        &=
        {\boldsymbol{y}}_t
        -
        {\boldsymbol{C}}_t
        \hat{\boldsymbol{x}}_{t|t-1},
        \\
        \boldsymbol{S}_t
        &=
        \boldsymbol{C}_t
        \boldsymbol{P}_{t|t-1}
        \boldsymbol{C}_t^T
        +
        \boldsymbol{R}_t,
        \\
        \boldsymbol{K}_t
        &=
        \boldsymbol{P}_{t|t-1}
        \boldsymbol{C}_t^T
        \boldsymbol{S}_t^{-1},
        \\
        \hat{\boldsymbol{x}}_{t|t}
        &=
        \hat{\boldsymbol{x}}_{t|t-1}
        +
        \boldsymbol{K}_t
        \tilde{\boldsymbol{y}}_t,
        \\
        \boldsymbol{P}_{t|t}
        &=
        \left(
        \I - 
        \boldsymbol{K}_t
        \boldsymbol{C}_t
        \right) 
        \boldsymbol{P}_{t|t-1},
    \end{align}
    with the innovation residual $\tilde{\boldsymbol{y}}_t$, the innovation covariance~$\boldsymbol{S}_t$, and the Kalman gain~$\boldsymbol{K}_t$.
\end{subequations}

\section{Collision Detection and Force Estimation}
To ensure the proposed algorithm remains independent of the number of feet, we assume that during the collision phase, the external force applied to one foot has negligible effects on the GM of the floating base and the other legs, similar to the approach in \cite{cha2023implementation}. This assumption is reasonable before a detrimental crash occurs, especially when an advanced balancing controller is deployed simultaneously. It is important to note that while the proposed algorithm can be applied to the full model, this would result in exponentially more contact modes and significantly higher computational complexity. Therefore, each leg uses an identical estimator but updates independently.
All the vectors with a subscript $j$ discussed later, unless otherwise specified, denote the segment corresponding to the $n_j$ DoF of $j^{\text{th}}$ leg.
\subsection{Mode-Dependent Process Model}
By combining the GM with a disturbance observer for contact forces \cite{wahrburg2015cartesian}, the mode-dependent system dynamics for the $j^{\text{th}}$ leg in a continuous format can be represented as
\begin{equation}
    \begin{bmatrix}
        \dot{\bm{p}}_j\\
        \dot{\bm{f}}_{\text{ext},j}
    \end{bmatrix}=\begin{bmatrix}
        \bolded{0} & \bolded{S}^{(k)} \bolded{J}_j^T \\
        \bolded{0} & \bolded{A}_f
    \end{bmatrix}\begin{bmatrix}
        \bolded{p}_j\\
        \bolded{f}_{\text{ext},j}
    \end{bmatrix} + 
    \begin{bmatrix}
        \bolded{I}\\
        \bolded{0}
    \end{bmatrix} \bolded{u} +
    \begin{bmatrix}
        \bolded{\omega}_p\\
        \bolded{\omega}_f
    \end{bmatrix},
\end{equation}
where the state is $\bolded{x}=[\bolded{p}_j^T,\bolded{f}_{\text{ext},j}^T]^T$, control is $\bolded{u}=(\bolded{\tau}_m - \bolded{\tau}_f + \bolded{C}^T\dot{\bolded{q}} - \bolded{g})_j$, and process noise is $\bolded{\omega} = [\bolded{\omega}_p^T, \bolded{\omega}_f^T]^T$. 
Here we define the jacobian $\bolded{J}_j \in \mathbb{R}^{3 \times n_j}$ by extracting the sub-block for $j^{\text{th}}$ leg from $\bolded{J}_{c_j}$. Note that we ignore external moments due to the point feet and the effects of the floating base following the earlier assumption.
The matrix $\bolded{A}_f \in \mathbb{R}^{6 \times 6}$, usually defined by a small negative definite diagonal matrix, governs the dynamics of the external force to mitigate the potential constant offset introduced by the disturbance \cite{yu2023fully}. The mode-dependent part is determined by the selection matrix $\bolded{S}^{(k)}$, where $k\in\{1,...,M\}$ denotes the $k^{\rm th}$ of the $M$ modes. In this work, we define $M=3$ with swing mode ($k=1$), stance mode ($k=2$), and collision mode ($k=3$). We define the selection matrix $\bolded{S}^{(k)}$,
\begin{equation}
    \bolded{S}^{(k)}=\begin{cases}
        \bolded{0} & \text{if} \quad k = 1\\
        \bolded{I} & \text{otherwise}
    \end{cases}.
\end{equation}
Therefore, in the process model, we differentiate only between the swing mode and other contact modes. For the swing mode, we assume zero external force, which is not factored into the prediction of the GM.

\subsection{Mode-Dependent Measurement Model}
In the measurement model, we introduce an additional \textit{pseudo-force} measurement to  better distinguish between different contact modes. The measurement model is
\begin{equation}
    \bolded{y} =
    \begin{bmatrix}
        \bolded{I} & \bolded{0}\\
        \bolded{0} & \bolded{I}
    \end{bmatrix}\begin{bmatrix}
        \bolded{p}_j\\
        \bolded{f}_{\text{ext},j}
    \end{bmatrix} + \begin{bmatrix}
        \bolded{v}_{p}\\
        \bolded{v}_{f}
    \end{bmatrix},
\end{equation}
where the output is the stack of the GM and contact force $\bolded{y}=[\bolded{p}_j^T,\bolded{f}_{\text{ext},j}^T]^T$ and the measurement noise $\bolded{v}=[\bolded{v}_p,\bolded{v}_f]^T$. The GM measurement is directly obtained from $\bolded{p} = \hat{\bolded{M}}\dot{\bolded{q}}$. Although a direct force measurement is not available due to the absence of a force sensor, a hypothetical \textit{pseudo force} can still be computed given the explicit contact mode. For the swing mode, we assume zero contact force. For the other contact modes, we combine the robot dynamics \eqref{eq:eom} and the contact constraint
\begin{equation}
    \bolded{J}_{c_j}\ddot{\bolded{q}} + \dot{\bm{J}}_{c_j}\dot{\bolded{q}} = 0.
\end{equation}
Hence, shorthanding $\bolded{\tau} = \bolded{\tau}_m - \bolded{\tau}_f - \bolded{g}$, the \textit{pseudo wrench} is 
\begin{equation}
    \bolded{\mathcal{F}}_{\text{pse},j} = -(\bolded{J}_{c_j}\bolded{M}^{-1}\bolded{J}_{c_j}^T)^\dagger(\bolded{J}_{c_j}\bolded{M}^{-1}\bolded{\tau}+\dot{\bm{J}}_{c_j}\dot{\bolded{q}}).
\end{equation}
where $\bolded{\mathcal{F}}_{\text{pse}} = [\bolded{f}_{\text{pse}}^T, \bolded{m}_{\text{pse}}^T]^T$ and $(\cdot)^\dagger$ denotes the pseudo-inverse. Assuming the joint velocity is zero during contact, a further simplified expression 
 is \cite{focchi2013local,menner2024simultaneous}
\begin{equation}\label{eq:simplified_force_cal}
    \bolded{\mathcal{F}}_{\text{pse},j} = -(\bolded{J}_{c_j}^T)^\dagger \bolded{\tau}_m.
\end{equation}
Instead of directly using \eqref{eq:simplified_force_cal}, our explicit definition of multiple contact models allows us to use this \textit{pseudo force} as an additional feedback for noncontact (swing) mode and other contact modes separately. Again, in this work we are only using $\bolded{f}_{\text{pse}}$ due to the point contact.

\begin{figure}
    \centering
    \includegraphics[width=0.9\linewidth]{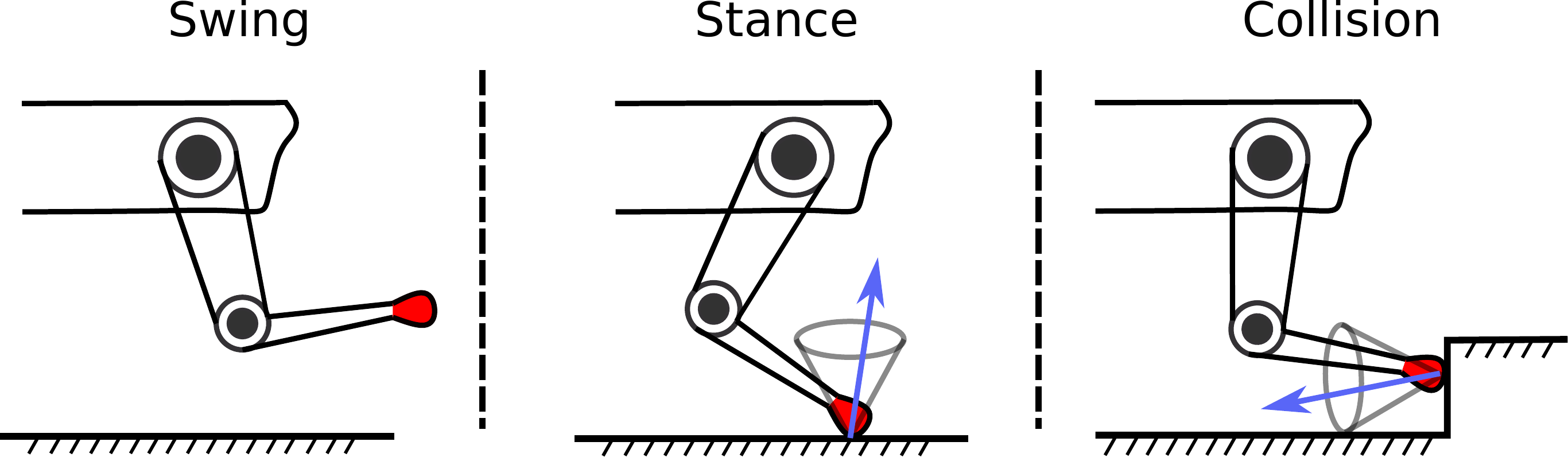}
    \caption{Force measurement models for different contact modes.}
    \label{fig:measurement_model}
    \vspace{-0.15in}
\end{figure}

To further distinguish between the contact modes,
As shown in Fig.~\ref{fig:measurement_model}, the shapes of the friction cones 
$\mathcal{F}^{(k)}$ are defined separately for the stance (vertical cone $\mathcal{F}_v$) and collision (horizontal cone $\mathcal{F}_h$) modes given prior assumption on the potential terrains. The measurement noise for the external force $\bolded{v}^{(k)}_f$ is then 
\begin{equation}
    \bolded{v}^{(k)}_f=\begin{cases}
        \text{large value} & \text{if} \quad \bolded{f}_{\text{pse},j} \in \mathcal{F}^{(k)}\\
        \text{small value} & \text{otherwise}
    \end{cases}
\end{equation}

\subsection{Interacting Multiple-Model Kalman Filter}
\label{ssec:immkf}
\begin{subequations}
Given the mode-dependent process and measurement models described above, we formulate an IMM-KF  for a switched system of the form
\begin{align}
    \boldsymbol{x}_{t+1}
    &=
    \boldsymbol A_t^{(k)} \boldsymbol{x}_t
    + 
    {\boldsymbol{B}}_{t}^{(k)}
    {\boldsymbol{u}}_{t} + \boldsymbol{w}_t^{(k)}
    \\
    \boldsymbol{y}_t
    &=
    \boldsymbol{C}_t^{(k)}\boldsymbol{x}_t
    + \boldsymbol{v}_t^{(k)}.
\end{align}
The switched system's modes are modeled as a Markov chain with transition probabilities
\begin{align}
    \prob
    \left(m_{t+1}=m^{(j)} \middle|
    m_{t}=m^{(i)}\right)
    =
    \pi_{ij},
\end{align}
where
$m_{t}\in\mathcal{M}$ is the mode at time step $t$, and the transition probability from mode~$m^{(i)}$ to mode~$m^{(j)}$ is $\pi_{ij}$.
The IMM-KF involves an interaction, a filtering, a probability update, and a combination step~\cite{blom1988interacting}.
\end{subequations}

\newcommand{\raisedMode}{{(k)}}
\subsubsection{Interaction step} 
\begin{subequations}
In the interaction step, the estimates of the $M$ filters are mixed and used to initialize each filter,
\begin{align}
    c^{(k)}
    &=
    \sum_{j=1}^{M}
    \pi_{jk} \mu_{t-1}^{(j)},
    \\
    \mu_{t-1|t-1}^{(j|k)}
    &=
    \frac{\pi_{jk}\mu_{t-1}^{(j)}}{c^{(k)}},
    \\
    \bar{\boldsymbol x}_{t-1|t-1}^{(k)}
    &=
    \sum_{j=1}^{M} 
    \mu_{t-1|t-1}^{(j|k)}
    \hat{\boldsymbol x}_{t-1|t-1}^{(j)},
    \\
    \bar{\boldsymbol{P}}_{t-1|t-1}^{(k)}
    &=
    \sum_{j=1}^{M} 
    \mu_{t-1|t-1}^{(j|k)}
    \left(
    \boldsymbol{P}_{t-1|t-1}^{(j)}
    +
    \boldsymbol{X}_{t-1|t-1}^{(k,j)}
    \right),
\end{align}
with $\boldsymbol{X}_{t|t}^{(k,j)}= \big( \bar{\boldsymbol{x}}_{t|t}^{(k)} - \hat{\boldsymbol{x}}_{t|t}^{(j)} \big)
\big( \bar{\boldsymbol{x}}_{t|t}^{(k)} - \hat{\boldsymbol{x}}_{t|t}^{(j)} \big)^T$,
where $\hat{\boldsymbol x}_{t|t}^{(k)}$ is the state estimate of filter~$k$ at time step $t$,
and
$\mu_{t}^{(k)}$ is the probability of filter~$k$ being active.

\subsubsection{Filtering step} 
Each Kalman filter is executed separately using  the mixed estimate $\bar{\boldsymbol x}_{t-1|t-1}^{(k)}$ and $\bar{\boldsymbol{P}}_{t-1|t-1}^{(k)}$ to initialize filter~$k$.

\subsubsection{Probability update step} 
The filters innovation residuals are used to update the filters model probabilities,
\begin{align}
\label{eq:mode_likelihoods}
    L_t^\raisedMode
    &=
    \frac{
    \exp{\left(
    -\frac{1}{2} 
     \big(\tilde{\boldsymbol y}_t^\raisedMode\big)^T 
    \big(\boldsymbol S_t^\raisedMode \big)^{-1} 
    \tilde{\boldsymbol y}_t^\raisedMode
    \right)
    }}{
    \left|2\pi \boldsymbol S_t^{(k)}\right|^{0.5}
    }
    \\
    \mu_t^\raisedMode
    &=
    \frac{c^\raisedMode L_t^\raisedMode}{
    \sum_{i=1}^M c^{(i)} L_t^{(i)}
    }
\end{align}
with the likelihood $L_t^\raisedMode$ and the model probability $\mu_t^\raisedMode$.

\subsubsection{Combination step} 
Lastly, the filters' state estimates are combined as a weighted sum using the model probabilities,
\begin{align}
    \hat{\boldsymbol{x}}_{t|t}
    &=
    \sum_{k=1}^M
    \mu_t^\raisedMode \hat{\boldsymbol{x}}_{t|t}^\raisedMode
    \\
    {\boldsymbol{P}}_{t|t}
    &=
    \sum_{k=1}^M
    \mu_t^\raisedMode 
    \left(
    {\boldsymbol{P}}_{t|t}^\raisedMode
    +
    \big(
     \hat{\boldsymbol{x}}_{t|t}
     - 
     \hat{\boldsymbol{x}}_{t|t}^\raisedMode
    \big)
    \big(
     \hat{\boldsymbol{x}}_{t|t}
     - 
     \hat{\boldsymbol{x}}_{t|t}^\raisedMode
    \big)^T
    \right),
\end{align}
\end{subequations}
which is the minimum mean-square estimate of the IMM-KF. Hence, this provides  the estimated external force $\hat{\boldsymbol{x}}_{t|t}$ and the probability $\mu^{(k)}_t$ of each contact mode being active.

\section{Collision-Aware Motion Control}
Although the design of contact-aware motion control algorithms remains an active field \cite{pang2022easing}, we demonstrate the advantages of accurate collision detection and external force estimation by integrating this information into a widely adopted locomotion control framework \cite{di2018dynamic}. The collision-aware adjustments include a reflex motion design, a swing leg Admittance Control (AC) law, and an MPC with external force feedback.

\subsection{Reflex Motion Design}
Upon detecting a collision on one leg, a reflex motion is triggered. If the swing foot is in the first half of its swing duration, as determined by the gait scheduler, the original step height is increased by a fixed amount  above the current foot height at the moment of collision. In our validations we use 10 cm increase of the foot height, but this can be practically adjusted as needed. The desired swing foot trajectories in the forward ($x$) and lateral ($y$) directions remain almost unchanged and only follow a newly interpolated trajectory towards the original foothold to further evaluate the swing leg AC, which is discussed next.

\subsection{Swing Leg Admittance Control}
We deploy an AC scheme to react to the external force,
\begin{equation}
    \ddot{r} = M_a^{-1} (\hat{f}_{\text{ext}} - D_a (\dot{r} - \dot{r}_d) - K_a (r - r_{d})) + \ddot{r}_{d},
\end{equation}
where the scalars $r$ and $r_d$ are the current and desired foot positions in the $x$ or $y$ direction.  The estimated external force $\hat{f}_{\text{ext}}$ is applied only after the collision event is detected. The parameters $M_a$, $D_a$, and $K_a$ are the virtual mass, damping coefficient, and stiffness coefficient, which can be tuned based on the practical system. Compared to a traditional operational space control (OSC) used in \cite{di2018dynamic, schperberg2022auto}, the inclusion of estimated external forces enables a compliant reaction to unexpected collisions, minimizing the impact force while compromising the original reference trajectory tracking error.
\subsection{MPC with External Force Feedback}
In this work, we use a convex PC \cite{di2018dynamic} based on a single rigid body model. Unlike the original MPC implementation in \cite{di2018dynamic}, when a collision is detected, we incorporate the estimated external force applied to the swing leg into the dynamics of the single rigid body model. This allows the MPC to account for disturbances caused by unexpected collisions, thereby enhancing the robot's balancing capabilities.

\section{Results}
We verify the proposed approach through both simulation and hardware experiments on a quadrupedal robot Unitree A1. The essential kinematic and dynamic terms are efficiently calculated using Pinocchio \cite{carpentier2019pinocchio}.

\subsection{Estimator Setup}
Table \ref{tab:estimator_param} shows the estimator parameters for our method. The transition probability matrix (TPM) $\bolded{\pi}$ is chosen as:
\begin{equation*}
    \bolded{\pi} = \begin{bmatrix}
        \pi_1 & \frac{1-\pi_1}{2} & \frac{1-\pi_1}{2} \\
        1-\pi_2 & \pi_2 & 0 \\
        1-\pi_3 & 0 & \pi_3
    \end{bmatrix}
\end{equation*}
where $\pi_1=\pi_2=\pi_3=0.8$. This design is based on the observation that collision and stance modes typically do not transition directly from one to the other during movement.

\begin{table}[h]
    \centering
    \caption{Estimator Parameter Setup}
    \begin{tabular}{c c c c c c}
        \hline
       $\bolded{A}_{f}$ & $\bolded{\omega}_{p}$ & $\bolded{\omega}_{f}$ & $\bolded{v}_p$ & $\bolded{v}_f$  \\\hline
       -0.01 & 0.0001 & 10.0 & 0.0001 & 0.001 / 200 \\\hline
    \end{tabular}
    \label{tab:estimator_param}
\end{table}

\subsection{Gazebo Simulation and Benchmark}
\begin{figure}
    \centering
    \includegraphics[width=0.75\linewidth]{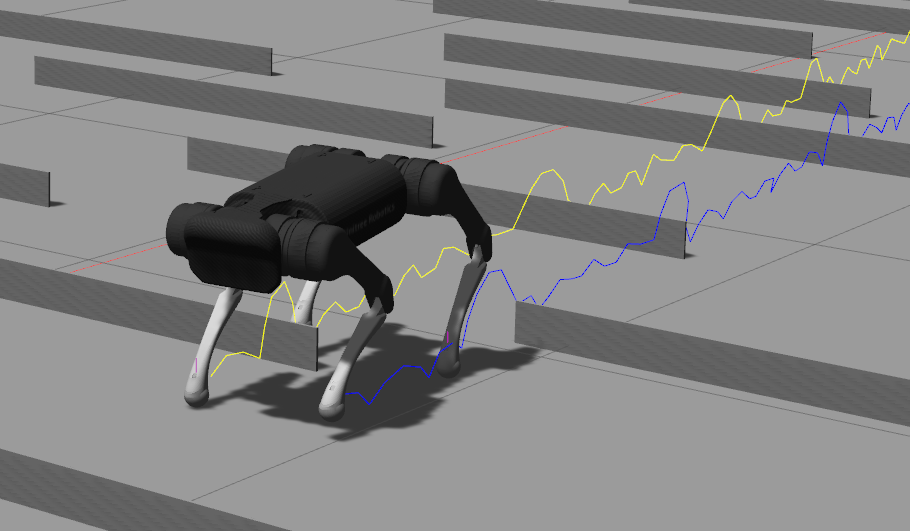}
    \caption{Gazebo simulation environment for collision tests.}
    \label{fig:gazebo_sim}
\end{figure}
Since the ground truth force measurements are absent on hardware, we first perform extensive studies in a Gazebo simulation engine. We establish an environment as shown in Fig.~\ref{fig:gazebo_sim}, where a set of rigid blocks are placed in front of the robot moving direction. The robot is commanded to trot at a speed of 0.5m/s. We compare  the following methods:
\begin{itemize}
    \item \textbf{FO-MBO}: first-order momentum-based observer (MBO) \cite{de2006collision}.
    \item \textbf{MBKO}: momentum-based Kalman observer (MBKO) through a combination of Kalman filter and disturbance observer \cite{wahrburg2015cartesian}.
    \item \textbf{PM-MBKO}: pseudo measurements (\ref{eq:simplified_force_cal}) MBKO using a single Kalman filter.
    \item \textbf{IMM-MBKO}: our proposed IMM-KF based MBKO.
\end{itemize}
For the methods we compare with, since the likelihood of different contact modes is not explicitly computed, we use a similar logic as in Fig.~\ref{fig:measurement_model} to determine the contact mode based on the estimated force distribution. Thresholds and estimator parameters are separately tuned to achieve best possible performance. We exclude comparisons with other heuristics-based approaches that require inputs from a planner or controller, as these are dependent on specific controller designs and extensive tuning.

\begin{figure}
    \centering
    \includegraphics[width=\linewidth]{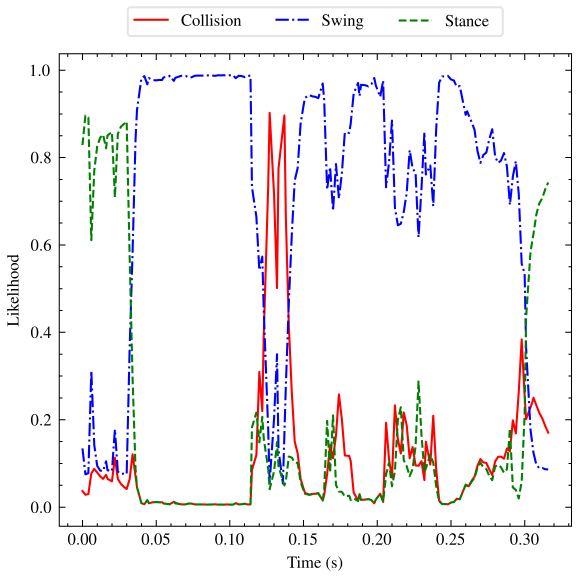}
    \vspace{-0.25in}
    \caption{The likelihoods of different contact modes estimated by the IMM-MBKO. The collision likelihood starts ramping up when a collision happens at around 0.11s.}
    \label{fig:likelihood_sim}
    \vspace{-0.15in}
\end{figure}

\begin{figure}
    \centering
    \includegraphics[width=\linewidth]{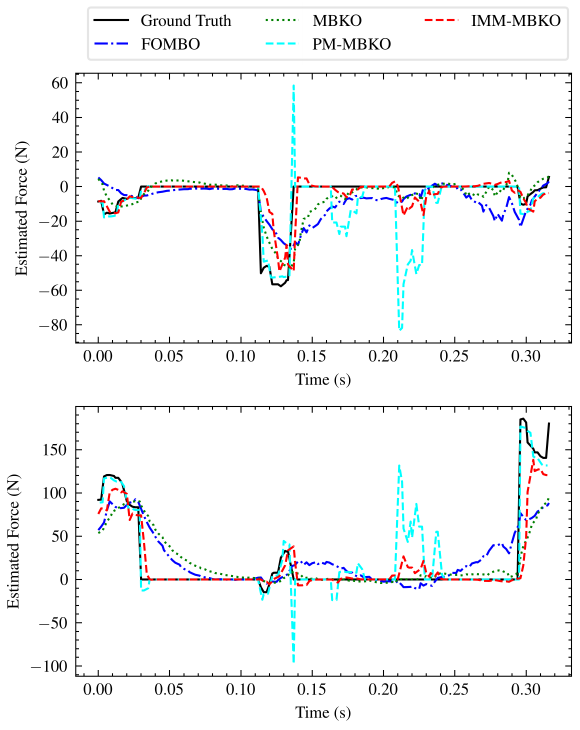}
    \vspace{-0.25in}
    \caption{Estimated forces in $x$ (upper) and $z$ (lower) directions.}
    \label{fig:estimated_force_sim}
\end{figure}

Table \ref{tab:benchmark_results} presents the collision-detection results after 89 test collisions, including the number of successful detections, detection delay, and instances of false detection (false positives for early detections and false negatives for missed detections). To further evaluate the quality of the estimated external forces, we include additional metrics: the absolute error in the magnitude of the collision force, as well as the swing and post-collision force Root Mean Square Error (RMSE) compared to the ground truth data. Following the approach in \cite{van2022collision}, we determine the absolute error \begin{equation}\label{eq:abs_force_error}
    e = \bigg|\biggl(\frac{\|\bolded{\hat{F}^{*}}\|}{\|\bolded{F^{*}}\|}-1\biggl) \cdot 100\%\bigg|
\end{equation}  to reflect the magnitude of estimated forces during collisions across all directions. These metrics including \eqref{eq:abs_force_error} provide a comprehensive analysis of the accuracy of force estimation before, during, and after collision events.

\begin{table}[t]
    \caption{Observer Benchmark Results}
    \centering
     \begin{tabular}{c c c c c}
         \hline
         & FO-MBO & MBKO & PM-MBKO & IMM-MBKO\\\hline
         Success/Total & \color{red}{74}\color{black}{/89} & 84/89 & 84/89 & \color{ForestGreen}{85}\color{black}{/89}\\
         False positive & \color{red}{69} & 4 & 10 & \color{ForestGreen}{1} \\
         False negative & \color{red}{9} & 5 & \color{ForestGreen}{4} & \color{ForestGreen}{4} \\
         Delay (ms) & 15.08 & \color{ForestGreen}{13.44} & \color{red}{24.34} & 14.79\\
         Abs error (\%) & \color{red}{58.96} & 45.69 & \color{ForestGreen}{31.54} & \color{black}{33.09} 
         \\
         Swing RMSE (N) & 16.77 & \color{red}{17.25} & .56 & \color{ForestGreen}{4.46}
         \\\hline \begin{tabular}{c c}
              Post-collision\\
               RMSE (N)
         \end{tabular} &
         15.33 & \color{ForestGreen}{11.21} & \color{red}{25.15} & 12.43
         \\
         \hline
    \end{tabular}
    \vspace{-0.1in}
    \label{tab:benchmark_results}
\end{table}
IMM-MBKO outperforms the other methods  in most metrics. Metrics such as the magnitude of collision force error, detection delay, and post-collision RMSE are also comparable to the best-performing approach. This demonstrates that IMM-MBKO combines the benefits of using pseudo-measurements for improved force estimation accuracy and IMM-KF for distinguishing between contact modes.

Using a specific collision as a demonstration, Fig.~\ref{fig:likelihood_sim} illustrates how the likelihoods for different contact modes change when a collision occurs around 0.11 seconds. 
Fig.~\ref{fig:estimated_force_sim} shows the corresponding changes in the estimated forces on the $x$ and $z$ axes during the same time period. The IMM-MBKO shows a better match with the ground-truth force profile expanding the swing $\rightarrow$ collision $\rightarrow$ swing phases. It is worth noting that the PM-MBKO produces large estimated forces after the collision, primarily due to the high acceleration from the reflex motion. By using IMM-KF, the IMM-MBKO significantly reduces this inaccuracy by explicitly distinguishing between different contact modes.

\begin{figure}
    \centering
    \includegraphics[width=\linewidth]{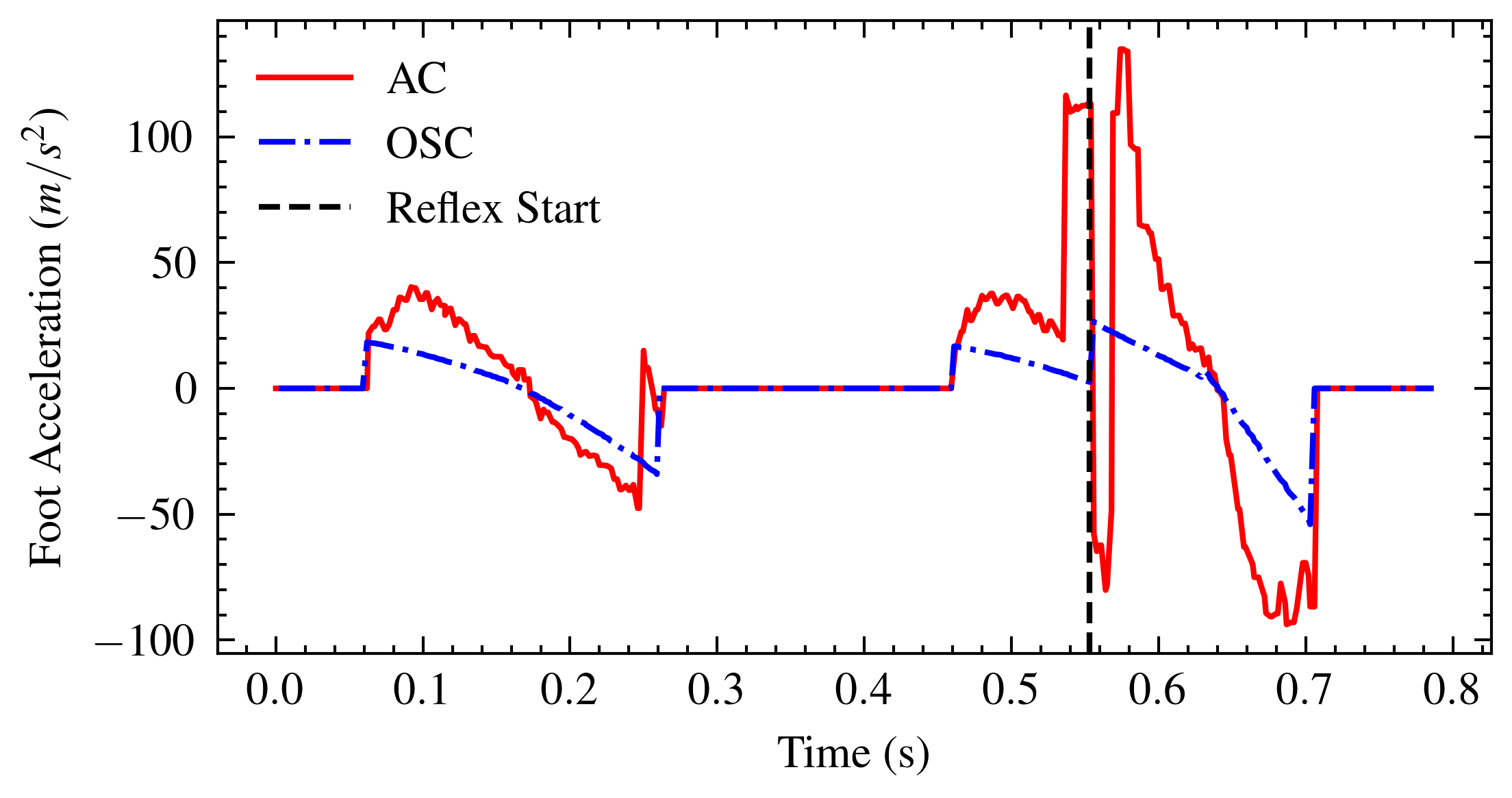}
    \vspace{-0.2in}
    \caption{Desired foot accelerations in the x direction for AC (red) and OSC (blue) before and after the reflex motion (black) is triggered.}
    \label{fig:adapted_acc}
\end{figure}

To demonstrate the proposed AC for the swing leg, Fig.~\ref{fig:adapted_acc} shows both how the AC and OSC modify the reference foot acceleration after a collision. The AC intentionally reacts to the estimated collision force by pulling the foot backward to reduce the impulse, while the OSC simply follows the newly interpolated trajectory. Table \ref{tab:ac_osc} presents a benchmark in the same simulation environment, comparing the performance of OSC with the original MPC in \cite{di2018dynamic} versus the proposed AC with MPC incorporating estimated force feedback. The results indicate significant reductions in base velocity tracking RMSE, average collision duration, and impulse.
\begin{table}[t]
    \caption{Swing Leg Control Schemes Comparisons}
    \centering
     \begin{tabular}{c | c | c}
         \hline
         & OSC + MPC w/o $\hat{f}_{\text{ext}}$ & AC + MPC w/ $\hat{f}_{\text{ext}}$\\\hline
         Total Collisions & 61 & 71\\
         Average Duration (s) & 0.052 & 0.036\\
         Velocity RMSE (m/s) & 0.090 & 0.072\\
         Average Impulse (Ns) & 2.134 & 1.15\\
         \hline
    \end{tabular}
    \vspace{-0.1in}
    \label{tab:ac_osc}
\end{table}
\subsection{Hardware Experiments}
To further validate the proposed algorithm on hardware, we tested the collision detection and reflex motion in a real-world scenario where a heavy table was placed in front of the robot. The table leg posed a potential tripping hazard that could cause damage. The step height was intentionally set lower than the table leg to verify the collision detection. The robot was operated at speeds ranging from 0.3--0.45m/s, with tests conducted for both forward and backward walking. The full performance can be viewed in the attached video.

\begin{figure}
    \centering
    \includegraphics[width=\linewidth]{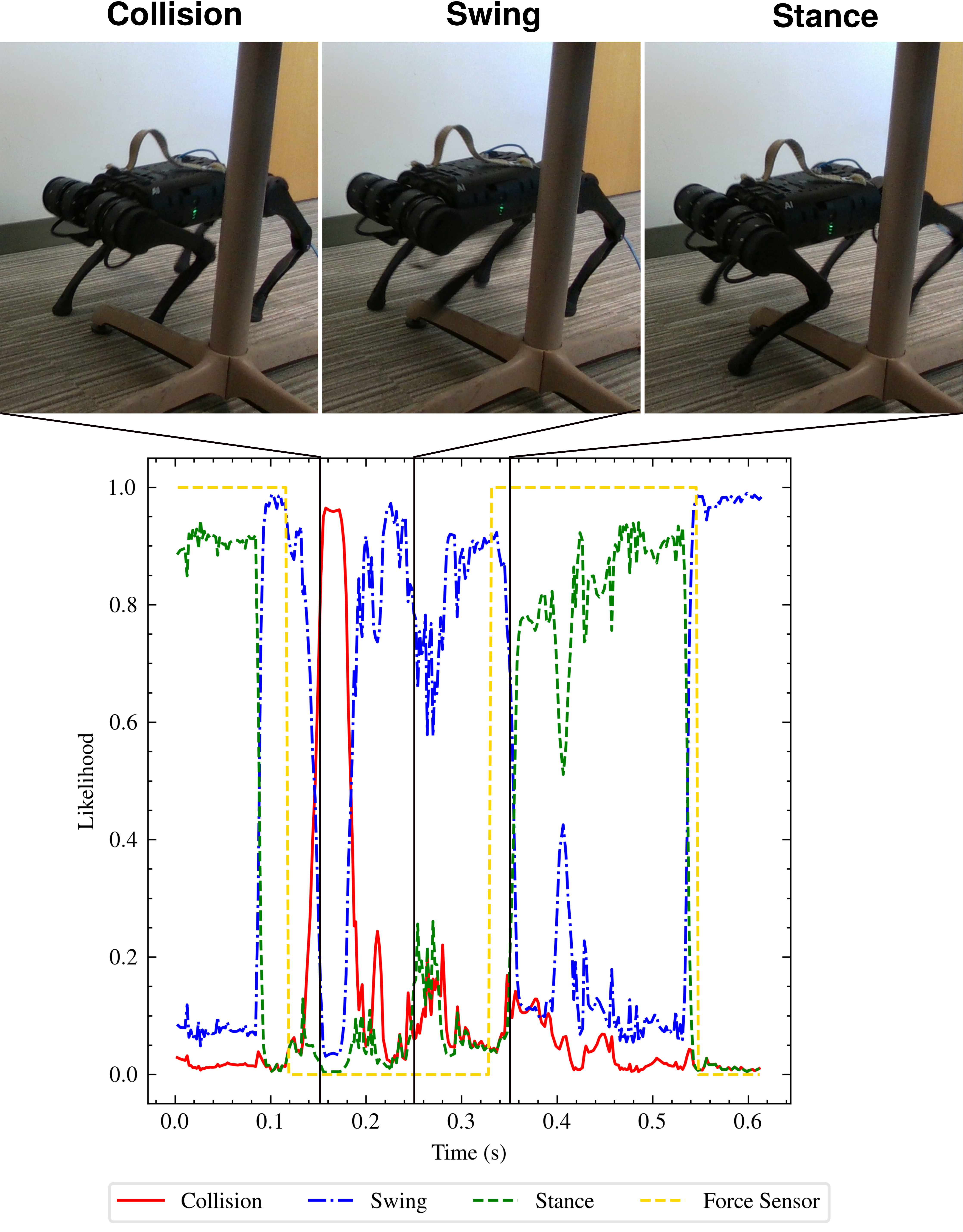}
    \caption{The likelihoods of each contact mode given by IMM-MBKO, along with the snapshots of the robot performing a reflex motion at different stages.}
    \label{fig:hardware}
    \vspace{-0.1in}
\end{figure}

Fig. \ref{fig:hardware} shows the hardware setup, with the front left foot performing the reflex motion, along with the corresponding contact likelihoods. Using a default low-resolution pressure sensor, we thresholded the sensor readings to mark the contact states. It was observed that the force sensor struggled to distinguish between collision and stance modes. In contrast, the IMM-MBKO detected the collision and informed the controller, preventing the robot from tipping over.

\subsection{Computational Time}
The IMM-MBKO for a single leg takes an average of 1.8ms, while the four-leg version takes 5.0ms, in a nonoptimized Python implementation, without any optimization for distributed implementation. The controller and estimators run on a computer equipped with an AMD 5950X processor.
 
\section{Conclusion}
This work presented a novel approach using an IMM-KF to simultaneously estimate external forces and contact modes, for enhanced robot collision detection and motion response. By integrating generalized momentum dynamics and external force estimation with pseudo measurements, the method provides reliable feedback for an AC and a force-adaptive MPC. The effectiveness is demonstrated through extensive simulations and hardware validations, highlighting its potential for robust, sensor-limited locomotion.

Several limitations were also identified in this study. First, the estimator may experience delays if the TPM is not properly tuned. Additionally, some rough prior knowledge of terrain shape is required to define the mode-dependent measurement models. Future work will focus on incorporating additional kinematic data, such as foot velocity, to further reduce detection delays. This information can be directly integrated into the TPM design. Moreover, integrating the proposed proprioceptive estimation approach with perceptive information could enhance accuracy. Lastly, the assumption of negligible effects on the floating base may limit the potential for more agile motions, which warrants further investigation.

\bibliographystyle{IEEEtran}
\bibliography{references}

\end{document}